\documentclass{article}

\usepackage{arxiv}

\usepackage[utf8]{inputenc} 
\usepackage[T1]{fontenc}    
\usepackage{hyperref}       
\usepackage{url}            
\usepackage{booktabs}       
\usepackage{amsfonts}       
\usepackage{nicefrac}       
\usepackage{microtype}      
\usepackage{lipsum}		
\usepackage{graphicx}
\usepackage{natbib}
\usepackage{doi}

\title{Adversarial Machine Learning in text analysis and generation}

\author{ \href{https://orcid.org/0000-0001-7832-5081}{\includegraphics[scale=0.06]{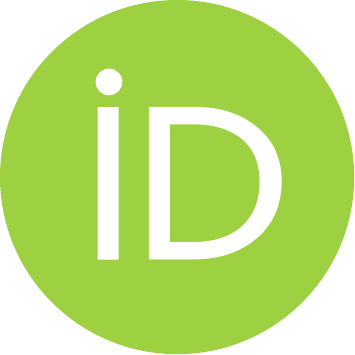}\hspace{1mm}Izzat Alsmadi}\\
	Department of Computing and Cyber Security\\
	Texas A\&M, San Antonio\\
	San Antonio, TX 78224 \\
	\texttt{ialsmadi@tamusa.edu} \\
}

\renewcommand{\shorttitle}{\textit{AML} in Text Analysis and Generation}

\hypersetup{
pdftitle={A template for the arxiv style},
pdfsubject={q-bio.NC, q-bio.QM},
pdfauthor={David S.~Hippocampus, Elias D.~Striatum},
pdfkeywords={First keyword, Second keyword, More},
}

\begin{document}
\maketitle

\begin{abstract}
The research field of adversarial machine learning witnessed a significant interest in the last few years. A machine learner or model is secure if it can deliver main objectives with acceptable accuracy, efficiency, etc. while at the same time, it can resist different types and/or attempts of adversarial attacks. This paper focuses on studying aspects and research trends in adversarial machine learning specifically in text analysis and generation. The paper summarizes main research trends in the field such as GAN algorithms, models, types of attacks,  and defense against those attacks. 

\end{abstract}


\section{Introduction}

A basic Generative Adversarial Network (GAN) model includes two main modules, a generator, and a discriminator. The generator and discriminator are implicit function expressions, usually implemented by deep neural networks \cite{creswell2018generative}.

Applying GAN in Natural Language Processing (NLP) tasks such as text generation is challenging due to the discrete nature of the text. Consequently, it is not straightforward to pass the gradients through the discrete output words of the generator \cite{haidar2019textkd}.

As text input is discrete, text generators model the problem as a sequential decision making process. In the model, the state is the previously generated characters, words, or sentences. The action or prediction to make is the next character/word/sentence to be generated. The generative net is a stochastic policy that maps current state to a distribution over the action space.

A generative adversarial network GAN can create new data instances that resemble original training data. Original GAN described by \cite{goodfellow2014explaining} has the following components and workflow:

\begin{itemize}
	\item Two NNs: a discriminator and a generator. The discriminator role is as of a simple classifier, that should distinguish real instances (positive examples, from the original training dataset) from fake instances (i.e. negative example), created by the generator.
	\item	The generator tries to fool the discriminator by synthesizing fake instances that resemble real ones. As training progresses, the generator gets closer to producing instances that can fool the discriminator. If generator trained very well, the discriminator gets worse at telling the difference between real and fake. Generally speaking, the generator module task is harder than that of the discriminator. For the least, the discriminator job is binary, but the generator job is much more complex.
	\item The two NNs compete with two different goals. The goal of the discriminator is to discriminate between the real and the fake instances. The goal of the generator is to eventually learn more about the real instances or data and fool the discriminator.
	\item	Both are trained separately. Each one assumes that the other module is fixed at the time of cycle training (to avoid dealing with a moving target that can be more complex). An accuracy of 100\% for the generator to generate fake instances indicates an accuracy of 50\% for the discriminator.
	\item	As another sign of competition/game between rivals, the generator instances become negative training examples for the discriminator. The discriminator punishes the generator for producing incorrect instances and rewards it for producing correct instances. The generator evolves to make the discriminator punishes less and rewards more.
	\item	In effect, a good discriminator should not reveal enough information for the generator to make progress.
	In addition to discriminator and generator modules, GANs include the following components:
	\item	Generator random input module
	\item	A generator network that transforms the random input into a data instance
	\item	Generator loss that punishes the generator for failing to fool the discriminator
	\item	Back-propagation module: It adjusts weights by calculating the weight's impact on the output
	
\end{itemize}

\subsection{Discriminator/Generator: Conditional versus Joint Probability}

We can model the difference between the discriminator and the generator as the difference between conditional probability p(Y | X) and joint probability p(X, Y) in given a set of data instances X and a set of labels Y. The conditional probability P(Y | X) is also known as the posterior probability for A. P(Y) and P(X) denote the prior probability of events Y and X, respectively. The conditional entropy indicates how much extra information you still need to supply on average to communicate Y given that the other party knows X. The joint entropy represents the amount of information needed to specify the value of two discrete random variables.

There are many variations on how each one of those two components (i.e. the generator and discriminator) work or is trained to function.

GANs have a problem in text generation since the gradients from discriminator can not be passed to the generator explicitly. To deal with this issue, GAN based models (e.g. SeqGAN: \cite{yu2016sequence}; Goal GAN \cite{florensa2018automatic},  MaliGAN \cite{che2017maximum} , LeakGAN \cite{guo2018long}; RankGAN \cite{lin2017adversarial}; MaskGAN \cite{fedus2018maskgan}; \cite{xu2018diversity}; \cite{caccia2018language} ) treat text generation as a sequential decision making process and utilize policy gradient; \cite{williams1992simple} to overcome this difficulty. The score predicted by a discriminator is used as the reinforcement to train the generator, yielding a hybrid model of GAN and RL. Other models utilize RL agents to control GANs. RL agent forms a decision-making network that interacts with the environment by taking available actions and collects rewards. As a scalability limitation, an agent that is trained using RL is only capable of achieving the single task that is specified via its reward function.

\subsection{GANs Methods for Training and Back-Propagation}

Applying GAN in text analysis is challenging as text is discrete. Consequently, their is a need to pass the gradients through the discrete output words of the generator, \cite{haidar2019textkd}.But how do generator and discriminator modules, train themselves to improve ?\\ 
The generator module can train itself using self, auto or variational encoders. The code in the decoder can randomly take values to produce different outputs. The goal of the self-encoder is to make the reconstruction error smaller and smaller. The generator uses back-propagation from the discriminator to improve its future instances and update model weights. 

The discriminator can also train itself, learn a discriminator process, that can distinguish an instance label as real or generated. It is trained based on real instances from the original dataset and fake instances from the generator. 

\subsection{GAN loss functions}

GAN uses loss functions that evaluate the distance between the distribution of the data that is generated by the GAN and the distribution of the real data. A GAN can have two loss functions for generator and discriminator training.

\begin{itemize}
	\item Minimax loss: The generator tries to minimize the loss function while the discriminator tries to maximize it, Goodfellow et al 2014
	\item Wasserstein loss: The default loss function for TF-GAN Estimators, \cite{arjovsky2017wasserstein}. In those GANs, the discriminator will not try to make a binary decision whether an instance is real or fake, but to provide a value between zero and one. A threshold can be decided in this range between values for fake instances versus values for real instances.
	
\end{itemize}
\section{Character/word/sentence level attacks }

While most of AML research publications demonstrate on picture-based datasets, a growing recent trend is the applications of AML in text analysis or NLP.
In one taxonomy, AML in NLP can be divided into the following attack levels:

\begin{itemize}
	\item Character-level attacks. Those involve different possible types of character-level manipulations such as: swap, substitution, deletion, insertion, repeating, one-hot character embedding, and visual character embedding (\cite{hosseini2017deceiving}, \cite{zhang2015character}, TextBugger \cite{li2018textbugger}, \cite{belinkov2017synthetic}, \cite{gao2018black}, Hotflip \cite{ebrahimi2017hotflip}, \cite{brown2019acoustic}, \cite{pruthi2019combating}, \cite{eger2019text}, \cite{le2020malcom}).
	
	While character-level attacks are simple, it is easy to defend against when deploying a spell check and proofread algorithms.
	
	\item Word-level attacks:	Similar to character-level attacks, approaches to word-level attacks through manipulation include:  
	word embedding, language models, filter words through synonyms. substitutes. (e.g., \cite{ebrahimi2017hotflip}, \cite{ebrahimi2018adversarial}, \cite{kuleshov2018adversarial}, 
	\cite{yang2020greedy},\cite{jin2020bert}, \cite{wallace2019universal}, \cite{gao2018black}, \cite{garg2020bae},\cite{zhou2019learning}, \cite{ribeiro2018semantically},\cite{zang2020word}, \cite{alzantot2019genattack}, \cite{li2016understanding}, \cite{li2018textbugger}, \cite{ribeiro2018semantically}, \cite{wang2019bilateral}, \cite{wang2020towards}).
	
	The search algorithms include gradient descent, genetic algorithms,  saliency-based greedy algorithm, sampling (\cite{papernot2016distillation}, \cite{sato2018interpretable}, \cite{gong2018adversarial}, \cite{alzantot2018generating}, \cite{zhou2019learning}, \cite{liang2017deep}, \cite{ren2019generating}, \cite{jin2020bert}).
	
	In comparison with character-level attacks, the attacks created by word-level approaches, are more imperceptible for humans and more difficult for machine learning algorithms to defend.
	
	\item Sentence-level attacks. Those attacks are usually based on text paraphrasing, demand longer time in adversary text generation. Examples of research publications in sentence level include (\cite{jia2017adversarial}, \cite{iyyer2018adversarial}, \cite{cheng2019robust}, \cite{michel2019evaluation}, \cite{lei2018discrete}, \cite{zheng2020understanding}, \cite{jethanandani2020adversarial}).
	\item Hybrid or multi-level attacks: Attacks which can use a combination of character, word, and sentence level approaches (HotFlip: \cite{ebrahimi2017hotflip}, \cite{blohm2018comparing}, \cite{wallace2019universal} ).
	
\end{itemize}

\section{Sequence generative/text generator/generative models approaches}

Natural Language Generation (NLG) techniques allow the generation of natural language text based on a given context.
NLG can involve text generation based on predefined grammar such as the Dada Engine, \cite{baki2017scaling} or leverage deep learning neural networks such as RNN, \cite{yao2017automated} for generating text. We will describe some of the popular approaches that can be found in relevant literature in the scope of AML.

\begin{itemize}
	\item Classical : training language models with teacher/professor forcing
	teacher forcing is common approach to training RNNs in order to maximize the likelihood of each token from the target sequences given previous tokens in the same sequence, \cite{williams1989learning}. In each time step, s, of training, the model is evaluated based on the likelihood of the target, t, given a groundtruth sequence. Teacher forcing is used for training the generator, which means that the decoder is exposed to the previous groundtruth token.\\ 
	RNNs trained by teacher forcing should be able to model a distribution that matches the target, where the joint distribution is modeled properly if RNN models prediction of future steps. Created error when using the model is propagated over each next or following step, resulting in low performance. A solution to this is training the model using professor forcing, \cite{lamb2016professor}.\\
	
	In professor forcing, RNN should give the same results when a ground truth is given as input (when training, teacher forcing) as when the output is looped back into the next step. This can be forced by training a discriminator that classifies wether the output is created with a teacher forced model or with a free running model, 
	
	\item	Conventional inference methods/ maximum likelihood estimation (MLE)\\
	 MLE is conducted on real data samples, and the parameters are updated directly according to the data samples. This may lead to an overly smooth generative model. The goal is to select the distribution that maximizes the likelihood of generating the data.
	For practical sample scenarios, MLE is prone to over-fitting/exposure bias issues on the training set. Additionally, during the inference or generation stage, the error at each time step will accumulate through the sentence generation process, \cite{ranzato2015sequence}\\.
	
	The following methods utilize MLE:
	\begin{itemize}
		\item 	Hidden Markov Model (HMM): 
		A Hidden Markov model (HMM) is a probability graph model that can depict the transition laws of hidden states, and mine the intentional features of data to model the observable variables. The foundation of an HMM is a Markov chain, which can be represented by a special weighted finite-state automaton. The majority of generative models require the utilization of Markov chains, \cite{goodfellow2020generative}, \cite{creswell2018generative}. The observable sequence in HMM is the participle of the given sentence in the part-of-speech PoS tag, while the hidden state is the different PoS.
		\item 	Method of moments: 
		The method of moments (MoM) or method of learned moments is an early principle of learning, \cite{pearson1893asymmetrical}. There are situations in which MoM is preferable to MLE. One is when MLE is more computationally challenging than MoM, Ravuri et al 2018.
		In the generalized method of moments (GMM), in addition to the data and the distribution class, a set of relevant feature functions is given over the instance space, \cite{hansen1982large}, \cite{rabiner1989tutorial}.
		Other research contributions in AML MoM or moment matching include: \cite{salimans2016improved}, \cite{mroueh2017fisher}, \cite{lewis2018adversarial}, \cite{bennett2019deep}.
		\item  Restricted Boltzmann Machine (RBM): 
		Restricted Boltzmann Machine (RBM) is a two-layer neural network consisting of a visible layer and a hidden layer, Hinton 2010. It is an important generative model that is capable of learning representations from data. 
		Generative models have evolved from RBM based models, such as Helmholtz machines (HMs), \cite{fodor1988connectionism} and Deep Belief Nets, DBN, \cite{hinton2006fast}, to Variational Auto-Encoders (VAEs), \cite{kingma2013auto} and Generative Adversarial Networks (GANs)
		
	\end{itemize}
	
	\item Cooperative training method 
	In Cooperative Training Method, CTM, a language model is trained online to offer a target distribution for minimizing the divergence between the real data distribution and the generated distribution, \cite{xie2017mitigating}, \cite{yin2020meta}.
	
	\item RL-based  versus RL-free text generation\\
	GAN models were originally developed for learning from a continuous, not discrete distribution. However, the 	discrete nature of text input handicaps the use of GANs
	
	In GANs, a reinforcement learning algorithm is used for policy gradient, to get an unbiased gradient estimator for the generator and obtain the reward from the discriminator, \cite{chen2018adversarial}.
	\begin{itemize}
		\item RL-based generation
		
		Reinforcement learning (RL) is a technique that can be used to train an agent to perform certain tasks. Due to its generality, reinforcement learning is studied in many disciplines. 
		
		GAN models that use a discriminating module to guide the training of the generative module as a reinforcement learning policy has shown promising results in text generation, \cite{guo2018long}. Various methods have been proposed in text generation via GAN (e.g. \cite{lin2017adversarial}, \cite{rajeswar2017adversarial}, \cite{che2017maximum}, \cite{yu2017seqgan}, \cite{che2017maximum}.

		There are several models of RL, some of which were applied to sentence generation, e.g., actor-critic algorithm and deep Q-network, (e.g. \cite{sutton2000policy}, \cite{guo2015generating}, \cite{bahdanau2016actor}. 
		
		One optimization challenge with RL-based approaches is that they may yield high-variance gradient estimates, \cite{maddison2016concrete}, \cite{zhang2017adversarial}.

		\item	RL free GANs for text generation\\
		Examples of models that use an alternative to RL:
		
		\begin{itemize}
			\item Latent space based solutions
			
			\item Continuous approximation of discrete sampling
		\end{itemize}
		
		 Those models apply a simple soft-argmax operator, or Gumbel-softmax trick to provide a continuous approximation of the discrete distribution on text.
		 
		 Examples of research efforts in this category include: 
		 TextGAN, \cite{zhang2017adversarial} and GumbelSoftmax GAN (GSGAN), \cite{kusner2016gans}, \cite{jang2016categorical}, \cite{maddison2016concrete},  FM-GAN, \cite{chen2018adversarial}, GSGAN, \cite{kusner2016gans}, and RelGAN, \cite{nie2018relgan}.
	\end{itemize}

\end{itemize}
\subsection{Long versus short text generation}
Literature in this area differentiates between the generation of short texts (e.g. less than 20 words) and the generation of long text. Applications for each one can be different from the other.

The majority of publications focus on short text generation as it seems to be less challenging. Different challenges are discussed in the literature specially in long text generation. For example, one of the unique challenges for long text generation is the sparse reward issue, in which a scalar guiding signal is only available after an entire sequence has been generated. \cite{vezhnevets2017feudal}, \cite{guo2018long}, \cite{sutton2000policy}. The main disadvantage
of sparse reward problem is making the training sample inefficient, \cite{tuan2019improving}.  
Model-based RLs have been proposed recently to solve problems with extremely sparse rewards, \cite{pathak2017curiosity}.

\subsection{Supervised versus unsupervised text generation}

Majority of work in this area falls in the supervised category (e.g. \cite{robin1994revision},\cite{tanaka1998reactive}, \cite{bahdanau2016actor},  \cite{bengio2015scheduled}, \cite{vinyals2015neural}, \cite{wiseman2017challenges}, \cite{bhowmik2018generating}, \cite{puduppully2019data}).\\

As a supervised problem, in a particular sentence, the terms/words in the sentence can be seen as the input features while the next term/feature is the target.

Examples of publications that fall in the unsupervised text generation include: ( \cite{graves2013generating}, \cite{yu2017seqgan}, \cite{zhang2018single}, \cite{hu2017controllable}, \cite{schmitt2020unsupervised}). Unsupervised text can be generated from explainable latent Topics, \cite{wang2018topicgan}, structured data, \cite{schmitt2019unsupervised}, \cite{sheffergoing}, or Knowledge graphs (KGs), \cite{bhowmik2018generating}, \cite{koncel2019text}, \cite{schmitt2020unsupervised}, \cite{jin2020genwiki}.

\section{Machine learning algorithms for text generation}
\begin{itemize}
	\item Using RNN (LSTM versus GRU versus BidirectionalRNN) for text generation\\
	
	State-of-the-art text-generation models are based on recurrent neural networks (RNNs). Several papers discussed using different deep learning RNN algorithms such as those mentioned above in automatic text generation, (e.g. \cite{kiddon2016globally}, \cite{hu2018texar}, \cite{abdelwahab2018deep}, \cite{lu2018neural}, \cite{nie2018relgan}, \cite{zhu2018texygen}, \cite{wang2019data}, \cite{mangal2019lstm}, \cite{moila2020development},  	\cite{mangal2019lstm}). Unlike traditional methods, RNN-based approaches rely on data-driven without manual intervention and emphasize on end-to-end encoder-decoder structure.\\
	Different performance metrics and methods are used to evaluate the output of the process such as log-likelihood, loss function, overall processing time, etc. The loss function that is used when training the model is the negative log likelihood or the negative log probability on the target sequence.\\ 
	LSTM shows to be a very good model in several aspects in comparison with the other evaluated models, \cite{mei2015talk},\cite{zang2017towards}, \cite{mangal2019lstm}. Many recent GANs for text generation, such as, \cite{kusner2016gans}, \cite{yu2017seqgan}, \cite{guo2018long}, \cite{lin2017adversarial} and \cite{fedus2018maskgan} are using LSTM. \\
	 
	 Some papers such as \cite{sutskever2011generating} and \cite{pouget2014overcoming} have shown that standard LSTM
	decoder does not perform well in generating long text sequences.
	\item Template-based, Rule-based versus neural text generation\\
	
	Classical approaches to text generation include: template-based, rule-based, n-gram-based and log-linear based models. Rule-based techniques are grammar-based methods with structured-rules written based on accumulated knowledge. Template-based approaches can be as simple as replacing words of users' choices by their synonyms, \cite{reiter1995nlg}, \cite{deemter2005real}, \cite{wiseman2018learning}, \cite{peng2019text}.\\
	
	N-gram models are widely used in NLP tasks such as text generation. In n-gram approach, the last word of the n-gram (i.e. to be predicted) can be inferred from the other words that already appear in the same n-gram, \cite{de2010improved}. 
	\item Beam search and Greedy Search\\
	Two popular deterministic decoding approaches are beem search
	and greedy search, \cite{sutskever2014sequence}, \cite{bahdanau2014neural}, \cite{zheng2019speculative}.  Beam search maintains 	a fixed-size set of partially-decoded sequences. Beam search is a common search strategy to improve results for several tasks such as text generation, machine translation and dependency parsing.  Greedy search selects
	the highest probability token at each time step. Greedy search can be seen as a special case of beam search.
	\item Sequence to sequence models and knowledge enhancement methods\\
	 Seq-to-Seq models are common architectures for text generation tasks where both the input and the output are modeled as sequences of tokens. In other words, the model convert an input sequence into an output sequence. More specifically, the first model encodes the input sequence as a set of vector representations using a recurrent neural network (RNN). The second RNN then decodes the output sequence step-by-step. Seq-to-Seq models are commonly trained via maximum likelihood estimation (MLE), \cite{chen2019improving}.
	 
	 One challenge with seq-to-seq models is that the input text alone often does not provide enough knowledge to generate the desired output which will impact the quality of the generated output. Several methods are proposed to enhance model knowledge beyond input text such as attention, memory, linguistic features, graphs, pre-trained language models, and multi-task learning. Many of those techniques are listed in \cite{yu2020survey} and https://github.com/wyu97/KENLG-Reading. 
	 
	 One of those particular enhancement techniques is attention, \cite{bahdanau2014neural} in which an encoder compresses the input text and a decoder with an attention mechanism generates output target word(s). The decoder is bound to generate a sequence of text tokens.
	
	\item 	Recursive Transition Network (RTN)\\
	The authors in \cite{baki2017scaling} discuss using a Recursive
	Transition Network, \cite{woods1970transition} for generating fake content similar in nature to legitimate content. RTN is used to detect simplification constructs. Nodes of the graph are labeled, and arcs may be labeled with either node names or terminal symbols. RNNs are essentially equivalent to an extension of context-free grammars in which regular expressions are allowed on the right side of productions.

	\item	Relational memory\\
	The basic idea of relational memory is to consider
	a fixed set of memory slots and allow for interactions between memory slots through using self-attention mechanisms, \cite{vaswani2017attention}.  RM is proposed to record key information of 	the generation process, for example, record 
	the information from previous generation processes. The goal is to enhance the text generation process through such learning/memory as well as patterns for long text generation. Such RL can provide a stateful, rather than stateless text generation process. Self attention is also used between the memory slots to enable interaction between them and facilitate long term dependency modeling, \cite{vaswani2017attention}. 
	
	Several relational-based text generations that showed better ability of modeling longer-range dependencies are described in literature, \cite{santoro2018relational}, RelGAN, \cite{nie2018relgan}.

	\item	Google LM\\
	Released by Google, Google LM is a language pre-trained model that is trained on a billion-word corpus, a publicly available dataset containing mainly news data \cite{jozefowicz2016exploring}, \cite{chelba2013one}. It is based on a two-layer LSTM with 8192 units in each layer, \cite{garbacea2019judge}.

	\item	Scheduled Sampling (SS)\\
	 SS is proposed to bridge the gap between training and inference for sequence prediction tasks. It is used to avoid exposure bias in seq-to-seq generation, \cite{bengio2015scheduled}, \cite{mihaylova2019scheduled}.
	
	During the inference process of seq-to-seq generation, true previous
	target tokens are unavailable. As a result, they are thus replaced by tokens generated by the model itself, which may yield
	a discrepancy between how the model is used at training and inference, \cite{bengio2015scheduled}.\\
	One limitation with scheduled sampling is that target sequences can be incorrect in some steps since they are randomly selected from the ground truth data, regardless of how input was chosen, \cite{zheng2018ensemble}, \cite{ranzato2015sequence} (Ranzato et al., 2015).
	\item	Generating text with GANs\\
	GANs are implicit generative or Language Models (LMs) learned via a competition between a generator network and a discriminator network. The discriminator distinguishes uniquely GANs from other LMs. Particularly for our subject, AML, adversarial training with the discriminator is used in GANs as opposed to training based on solely maximum likelihood and categorical cross entropy in other LMs. The conventional LMs are not trained in a adversarial manner.
	Unlike traditional approaches (e.g. teacher forcing, SS), GANs do not suffer from exposure bias, \cite{rajeswar2017adversarial}, \cite{tevet2018evaluating}. Exposure bias occurs when models are fed with their predicted data rather than the ground-truth data at inference time. This causes generating poor samples due
	to the accumulated error, \cite{yin2020meta}.

\end{itemize}

\section{Adversarial training techniques}
Adversarial training is a method to help systems be more robust against adversarial attacks. Below are examples of some adversarial training techniques reported in literature.

\begin{itemize}
	\item Fast Gradient Sign Method FGSM\\
	FGSM is used to add adversarial examples to the training
	process \cite{goodfellow2014explaining}, \cite{wong2020fast}. During training, part of the original samples is replaced with its corresponding adversarial samples generated using the model being trained.
	
	Kurakin et al. suggested to use Iterative FGSM, IFGSM, FGSM-LL or FGSM-Rand variants for adversarial training, in order to reduce the effect of label leaking, \cite{kurakin2016adversarial}. Their are also other variants of FGSM such as: Momentum Iterative Fast Gradient Sign Method (MI-FGSM), \cite{dong2018boosting}.
	
	\item PGD-based training\\
	Proposed by \cite{madry2017towards}. At each iteration all the
	original samples are replaced with their corresponding adversarial samples generated using the model being trained.
	
	PGD was enhanced using different efforts such as:  
	\begin{itemize}
		\item Optimization tricks such as momentum to improve adversary, \cite{dong2018boosting}
		\item Combination with other heuristic defenses such as matrix estimation, \cite{yang2019me}
		\item Defensive Quantization, \cite{lin2019defensive}
		\item Logit pairing, \cite{mosbach2018logit}, \cite{kannan2018adversarial}
		\item Thermometer Encoding, \cite{buckman2018thermometer}
		\item Feature Denoising, \cite{xie2019feature}
		
		\item  Robust Manifold Defense, \cite{jalal2017robust}
		\item L2 nonexpansive nets, \cite{qian2018assigning}
		\item Jacobian Regularization, \cite{jakubovitz2018improving}
		\item Universal Perturbation, \cite{shafahi2020universal}
		\item Stochastic Activation Pruning, \cite{dhillon2018stochastic} 
		
	\end{itemize}
	
	As of today, training with a PGD adversary remains empirically robust, \cite{wong2020fast}
	\item Jacobian-based saliency map approach (JSMA)\\
	
	JSMA is a gradient based white-box method that is proposed to use the gradient of loss with each class labels with respect to every component of the input, \cite{papernot2016limitations}.
	JSMA is useful for targeted miss-classification attacks, \cite{chakraborty2018adversarial}.

	\item Accelerating Adversarial Training\\
	The cost of adversarial training can be reduced by reusing adversarial examples and merging the inner loop of a PGD and gradient updates of the model parameters,  \cite{shafahi2019adversarial}, \cite{zhang2019you}
	
	\item DAWNBench competition\\
	
	Some submission projects to DAWNBench competition have shown good performance results on CIFAR10 and ImageNet classifiers in comparison with research-reported training methods, \cite{coleman2017dawnbench}, \cite{wong2020fast}.

\end{itemize}

\section{Text Generation Models/Tasks/Applications}
Text generation refers to the process of automatic or programmable generation of text with no or least of human intervention. The sources utilized for such generation process can also vary based on the nature of the application. The types of applications from generating text in particular are growing. We will discuss just a few in this section.

\subsection{Next-Word Prediction}
For many applications that we use through our smart phones, or websites, next word prediction (NWP, also called auto-completion) is a typical NLP application. From a machine-learning perspective, NWP is a classical prediction problem where previous and current text can be the pool to extract the prediction model features and other parameters and the next word to predict is the target feature. Different algorithms are proposed to approach NWP problem such as term frequencies, artificial intelligence, n-grams, neural networks, etc.
\subsection{Dialog Generation}

Human-machine dialog generation/prediction is an essential topic of research in the field of NLP. It has many different applications in different domains. The quality and the performance of the process can widely vary based on available resources, training/pre-training and also efficiency.

Seq2seq neural networks have demonstrated impressive results on dialog generation, \cite{vinyals2015neural}, \cite{chang2019semi}. GANs are used in dialogue generations in several research publications (e.g. \cite{li2016deep}, \cite{hamilton2017inductive}, \cite{kannan2018adversarial}, \cite{nabeel2019cas}, ) 

\subsection{Neural Machine Translation}

Neural Machine Translation (NMT) is a learning approach for automated translation, with potentials to overcome weaknesses of classical phrase-based translation systems or statistical machine learning. The main difference is that NMT is based on a model not based on some patterns. NMT tries to replicate the functions of the human brain and assess content from various sources before generating output. Further enhancements on NMT were achieved using attention based neural machine translation. 

One of the popular early open source NMTs is Systran:  https://translate.systran.net/, the first NMT engine launched in 2016.Other examples include those of: Google Translate, Facebook, e-bay and Microsoft.

Adversarial NMT is introduced in which  training of the NMT model is assisted by an adversary, an elaborately designed 2D-convolutional neural network (CNN), \cite{yang2017improving}, \cite{wu2018adversarial}, \cite{zhang2018bidirectional}, \cite{shetty2018a4nt}.

\section {Text Generation Metrics}
One of the key issues in text generation is that there is no widely agreed-upon automated metric for evaluating the text generated output. Text generation metrics can be classified based on several categories. Here is a summary of categories and metrics:

\begin{itemize}
	\item Document Similarity based Metrics\\
	One of the popular approaches to measure output TG is through comparing it with some source documents or human natural language. Some of the popular metrics in this category are Bilingual Evaluation Understudy (BLEU), \cite{papineni2002bleu} and Embedding Similarity (EmbSim), \cite{zhu2018texygen}.
	
	 BLEU has several variants such as  BLEU-4 and BLEU-1.
	
	This category can also include some of the popular classical metrics such as:Okapi BM25 \cite{10.5555/188490.188561}, Word Mover’s Distance (WMD), \cite{10.5555/3045118.3045221}, Cosine,
	Dice and Jaccard measures in addition to Term Frequency-Inverse
	Document Frequency (TF-IDF). 
	
	\item Likelihood-based Metrics\\
	Log-likelihood is the negative of the training loss function, (NLL). NLL (also known as multiclass cross-entropy) outputs a probability for each class, rather than just the most likely class. The typical approach in text generation  is to train the model using
	a neural network performing maximum likelihood estimation (MLE)
	by minimizing the negative log-likelihood, NLL over the text corpus. For GANs,  in the standard GAN objective, the goal or objective function is to minimize NLL for the binary classification task, \cite{goodfellow2014explaining}. 
	
	Maximum Likelihood suffers from predicting most probable answers. This means that a model trained with maximum likelihood will tend to output short general answers that are very common in the vocabulary.
	The log-likelihood improves with more dimensions as it is easier to fit the hypotheses in the training step having more dimensions. Consequently, the hypothesis in the generating step have lower log-likelihood.
	\item Perplexity\\
	Perplexity measures a model’s certainty of its predictions.
	
	There are several advantages to using perplexity, \cite{keukeleire2020correspondence};
	\begin{itemize}
		\item Calculating perplexity is simple and doesn’t require human interference
		\item It is easy to interpret
		\item I is easy to optimize a model for an improved perplexity score
	\end{itemize} 
	Held-out likelihood is usually presented as perplexity, which is a deterministic transformation of the log-likelihood into an  information-theoretic quantity
	\item  Inception Score (IS)\\
	IS rewards high confidence class labels for each generated instance, \cite{salimans2016improved}. IS can provide a general evaluation of GANs trained on ImageNet. However, it has limited utility in other settings, \cite{fowl2020random}.
	\item Frechet Inception Distance (FID)\\
	FID is used to measure the Wasserstein-2 distance, \cite{vaserstein1969markov} between two Gaussians, whose 	means and covariances are taken from embedding 	both real and generated data, \cite{heusel2017gans}, \cite{cifka2018eval}. FID assumes that the training data is "sufficient" and does not reward producing more diversity than the training data , \cite{fowl2020random}.
	\item N-gram based metrics\\
	Distinct-n is a measure of diversity that computes the number of
	distinct n-grams, normalized by the number of all ngrams, \cite{li2015diversity}.  
	\item Sentence similarity metrics
	, SentenceBERT (sent-BERT, Reimers and Gurevych 2019)

\item ROUGE metrics\\
ROUGE metrics were mostly used for text generation, video captioning and summarization tasks , \cite{lin2004rouge}. They were introduced in 2004 as
a set of metrics to evaluate machine-generated text summaries. 
  ROUGE has several variants such as: ROUGE-1, ROUGE-2 and ROUGE-L.
 \item METEOR\\
 METEOR (Metric for Evaluation of Translation with Explicit Ordering) was proposed in 2005, \cite{banerjee2005meteor}. METEOR metric was mainly used for text generation, image and video captioning, and question answering tasks
\item Embedding-based metrics\\
The main approach is to to embed generated sentences in latent space and then evaluate them in this space, \cite{tevet2018evaluating}. \cite{du2019boosting} suggest to cluster the embedded sentences with k-means and then use its inertia as a measure for diversity. 
\item Other less common metrics such as: GLEU score, edit distance, phoneme and diacritic error rate.
\item Metrics for GANs, traditional probability-based LM metrics, \cite{tevet2018evaluating}
Several papers indicated the need to use new metrics to evaluate GANs (e.g. \cite{esteban2017real}, \cite{zhu2018texygen},\cite{saxena2019d}). 
Some of the metrics proposed for GANs include:
\begin{itemize}
	\item Divergence based Metrics such as F-GAN, \cite{nowozin2016f}, LS-GAN \cite{mao2017least}, KL-divergence \cite{koochali2019probabilistic},  and  Self-BLEU, \cite{zhu2018texygen}
	\item Integral Probability Metrics such as: Wasserstein GAN (WGA)' \cite{arjovsky2017wasserstein}, \cite{gulrajani2017improved}.
	\item Domain-specific metrics, e.g. attack success rate, \cite{gao2020generative}.
	
	\item Random Network Distillation (RND), \cite{burda2018exploration}.
\end{itemize}
\end{itemize}

\section{Text Generation Datasets}
There are many datasets for the general tasks/research of NLP such as those mentioned in the following links:

\begin{itemize}
	\item https://aclweb.org/aclwiki/Data\_sets\_for\_NLG
	\item https://paperswithcode.com/task/data-to-text-generation
	\item https://project-awesome.org/tokenmill/awesome-nlg\#datasets
	\item https://github.com/niderhoff/nlp-datasets
	\item https://lionbridge.ai/datasets/the-best-25-datasets-for-natural-language-processing/
	\item https://machinelearningmastery.com/datasets-natural-language-processing/
	\item https://www.kdnuggets.com/tag/datasets
\end{itemize}
 
 We will list few datasets/benchmarks that are used in GAN research papers in particular
 
 \begin{itemize}
 	\item COCO Image Captions, \cite{chen2015microsoft} and 
 	\item	EMNLP2017 WMT: http://statmt.org/wmt17/translation-task.html, \cite{guo2018long}
 	 
 	\item  WeiboDial, \cite{qian2018assigning}
 	\item	Chinese poems, a dataset proposed by \cite{zhang2014chinese} and used by other related work such as, \cite{yu2017seqgan}, \cite{lin2017adversarial}, \cite{rajeswar2017adversarial}.
 	\item	CUB captions, \cite{wah2011caltech}
 	\item	Dada Engine to create phishing emails, \cite{baki2017scaling}
 	
 \end{itemize}

\section{Memory based models, RNN versus LSTM}
 
 As we mentioned earlier, vanilla RNNs do not perform well when the
 learning sequences have long term temporal dependence due
 to issues such as exploding gradients, \cite{bengio2015scheduled}.
 Alternatively, Convolutional neural networks (CNNs), recurrent neural networks (RNNs), Gated recurrent unit, (GRU) and Long-short
term memory (LSTM) models are effective approaches in the field
of sequential modeling methods. The design of the forget
gate is the essence of these models, \cite{sun2020graphpb}.

An LSTM model is a type of RNN that can remember
relevant information longer than a regular RNN. As a result, they can better learn long-term patterns  \cite{olah2015understanding}.

LSTM models provide a mechanism that is able to
both store and discard the information saved about the previous steps, limiting the accumulated error using Constant
Error Carousels, \cite{hochreiter1997long}, \cite{manzelli2018end}.

\section{Defense Against NLP Adversarial Attacks}

Generating adversarial attacks on text has shown to be
more challenging than for images and audios due to their discrete nature. Variations on original text can be applied on different levels, character, word or sentence levels. Recent relevant studies showed examples of NLP vulnerabilities such as, \cite{zhou2019learning}:

\begin{itemize}
	\item Reading comprehension, \cite{jia2017adversarial}.
	\item Text classification, \cite{alzantot2018generating}, \cite{liang2017deep}, \cite{wong2017dancin}
	\item Machine translation, \cite{cheng2020seq2sick}, \cite{ebrahimi2018adversarial}
	\item Dialogue systems, \cite{cheng2019robust}
	\item Dependency parsing, \cite{zheng2020understanding}. 
\end{itemize}

\subsection{Black versus white box attacks}
Current adversarial attacks can be roughly be divided into three categories: white-box attacks, black-box and gray-box attacks, according to whether the data, model architecture, and parameters of the target are accessible. In black-box attacks (also called zero-knowledge attack), no or very limited information about the target model is accessible. For example, a certain number of model queries (i.e. oracle queries) are granted.

Some of the defenses, \cite{guo2018long}, \cite{xie2017mitigating} are shown to be quite robust against black-box attacks.

In gray-box attacks/limited knowledge attacks, partial knowledge about the model under attack (e.g., type of features, or type of training data) is assumed.
On the other side, is white-box attack/ perfect-knowledge attacks. Those are attacks that exploit model internal information. They assume complete knowledge of the targeted model, including its parameter values, architecture, training method, and in some cases its training data. Table 1 shows samples of research publications in all three categories. There are some papers that are identified to more than one category.

At the high level, there are three classes or dimensions of attacks, \cite{barreno2010security}:

\begin{itemize}
	\item Causative versus Exploratory
	In causative attacks, the training process is altered and models are trained with adversary datasets. In exploratory attacks, attacker tries to exploit the existing weaknesses
	
	\item Integrity versus availability: false negatives versus false positive. 
		
	\item Targeted at a particular input or indiscriminate in which input fails
	
	\item Reactive versus proactive: A reactive defense is where one waits to be attacked and detects an adversarial example. On the other hand, proactive attacks involve training the model to be more resilient against adversarial 
	
\end{itemize}

Examples of Defense against Specific Attacks with focus on NLP:
\begin{itemize}
	\item  Dirichlet Neighborhood Ensemble (DNE), a randomized
	smoothing method for training a model against substitution-based attacks, \cite{zhou2020defense}
	
	\item Adversarial training as a defense method,\cite{miyato2016adversarial}, \cite{sato2018interpretable}, \cite{zhu2019freelb}.
	
	\item Increasing the model robustness by adding perturbations on word embedding, \cite{goodfellow2014explaining}
	
	\item Certified defenses: Some certified defenses were proposed in literature in order to provide guarantees of robustness to some specific
	types of attacks, \cite{huang2019achieving}, \cite{jia2017adversarial}. 
	
	\item Defensive distillation: Defensive
	distillation can take an arbitrary NN and increase its robustness, reducing the success rate of attacks’ ability, \cite{carlini2017towards}
	
	\item Defense through randomization, \cite{cohen2019certified}, \cite{liu2018towards}
	
\end{itemize}

\section{Summary and Conclusion}
In this paper, recent literature in adversarial machine learning for text generation tasks is summarized. Our goal is to present a one-stop source for researchers and interested readers to learn the basic components and research trends in this field. We noticed a continuous expansion in the applications, models and algorithms. This paper can serve as an introduction to this field and readers may need to follow through some of the researchers and references we referred to based on their focuses or interests.
 
\bibliographystyle{unsrtnat}
\bibliography{references}  






\end{document}